\documentclass[letterpaper]{article}
\usepackage[preprint]{aaai2027}
\usepackage[hyphens]{url}
\usepackage{graphicx}
\usepackage{natbib}
\usepackage{caption}
\usepackage{algorithm}
\usepackage{algorithmic}
\usepackage{booktabs}
\usepackage{colortbl}
\usepackage{listings}
\definecolor{tablegroupgray}{rgb}{0.9,0.9,0.9}
\usepackage{multirow}
\usepackage{amsmath,amssymb}
\usepackage{bm}
\usepackage{pifont}
\usepackage{xspace}

\newcommand{\method}{\textbf{EAVer}\xspace}
\newcommand{\searchtag}{\texttt{<search>}}

\newcommand{\memotag}{\texttt{<memo>}}
\newcommand{\verifytag}{\texttt{<verify>}}
\newcommand{\answertag}{\texttt{<answer>}}
\newcommand{\support}{\textsc{supported}\xspace}
\newcommand{\unsupport}{\textsc{unsupported}\xspace}
\newcommand{\fscore}{macro-F1\xspace}
\newcommand{\fsr}{FSR\xspace}

\title{\raisebox{-0.15\height}{\includegraphics[scale=0.045,trim=13bp 0 30bp 0,clip]{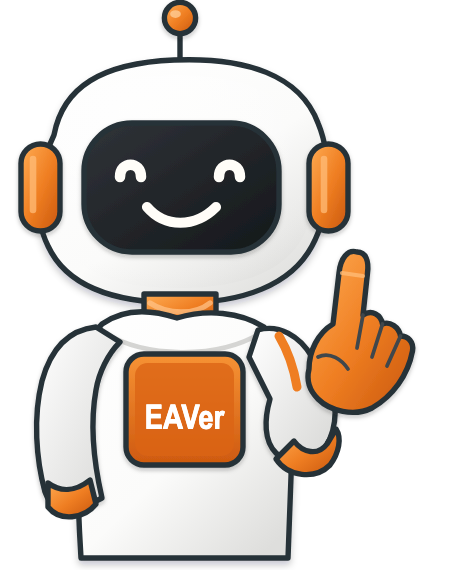}}\hspace{0.2em}\method: Long-Form Factuality Verification as an End-to-End Agentic Policy}
\author{
    \normalsize\mdseries
    \begin{tabular}{@{}c@{}}
        Kening Zheng\textsuperscript{1,2}\thanks{Work done during an internship.},
        Aoying Zheng\textsuperscript{2,3},
        Zhigang Chang\textsuperscript{2}, Yazhi Guo\textsuperscript{2},
        Miaotian Guo\textsuperscript{2}, Qingwei Zong\textsuperscript{2} \\[-0.08em]
        Xianhai Xie\textsuperscript{2}, Weiqiang Jin\textsuperscript{4},
        Chengze Li\textsuperscript{1}, Hanrong Zhang\textsuperscript{1},
        Jie Yang\textsuperscript{1}, Wei-Chieh Huang\textsuperscript{1} \\[-0.08em]
        Lingzhe Zhang\textsuperscript{1,5}, Liancheng Fang\textsuperscript{1},
        Xin Zou\textsuperscript{6}, Hanqian Li\textsuperscript{6},
        Jiahao Huo\textsuperscript{6}, Yibo Yan\textsuperscript{6} \\[-0.08em]
        Zizhuang Deng\textsuperscript{3}, Lei Miao\textsuperscript{2},
        Wei Guo\textsuperscript{2}, Haihong Tang\textsuperscript{2},
        Bo Zheng\textsuperscript{2}, Philip S. Yu\textsuperscript{1}\corresponding
    \end{tabular}
}
\affiliations{
    \small
    \begin{tabular}{@{}c@{}}
        \textsuperscript{1}University of Illinois Chicago
        \quad \textsuperscript{2}Taobao \& Tmall Group
        \quad \textsuperscript{3}Shandong University \\[-0.08em]
        \textsuperscript{4}Xi'an Jiaotong University
        \quad \textsuperscript{5}Peking University \\[-0.08em]
        \textsuperscript{6}The Hong Kong University of Science and Technology (Guangzhou)
    \end{tabular}
}

\begin{document}

\maketitle

\begin{abstract}
Long-form factuality verification is commonly implemented as a static decompose-search-verify pipeline, with separately prompted modules processing claims and invoking external search.
Treating claims independently makes LLM and search calls scale with claim count and causes repeated searches for overlapping evidence about related claims.
We introduce \method, an \textbf{E}nd-to-end \textbf{A}gentic \textbf{Ver}ifier that learns to control the complete response-level verification workflow as a unified policy.
\method groups semantically related claims, routes each group to direct verification or targeted search based on confidence, and keeps evidence returned by search in compact in-context memos for cross-claim reuse.
To train this policy, we develop a privileged-teacher synthesis pipeline that converts gold claim annotations into executable multi-turn tool-interaction trajectories with live search rather than post-hoc rationales.
Structural, label-alignment, tool-use, search-budget, and leakage checks yield 1{,}447 quality-controlled trajectories.
We further construct 794 bidirectional same-trajectory preference pairs that keep claim grouping, search, and evidence fixed, enabling decision-focused Direct Preference Optimization (DPO) over factuality-decision tokens.
The results with Qwen3-8B show that \method outperforms the strongest search-based baseline on each benchmark by 2.88 Macro-F1 points on VeriFastScore and 4.73 points on the out-of-distribution FaStFact-Bench, while using about 80\% fewer searches than the most search-efficient baseline.
Moreover, \method consistently improves performance across models ranging from 4B to 32B parameters, demonstrating its strong generalizability.
\end{abstract}

\section{Introduction}

Large language models (LLMs) can produce fluent, extended responses across open-ended question answering and instruction-following settings~\citep{brown2020language,ouyang2022training}.
Yet fluency does not guarantee factual reliability: generated answers may contain plausible but unsupported statements that are difficult to detect at the response level~\citep{lin2022truthfulqa,manakul2023selfcheckgpt}.
This challenge is amplified in long-form generation, where factual statements are distributed across sentences and paragraphs and may connect entities, dates, quantities, events, and causal relations.
Reliable evaluation therefore requires fine-grained verification that identifies individual claims and grounds each judgment in the relevant evidence for that claim.

\begin{figure}[t]
\centering
\includegraphics[width=\columnwidth]{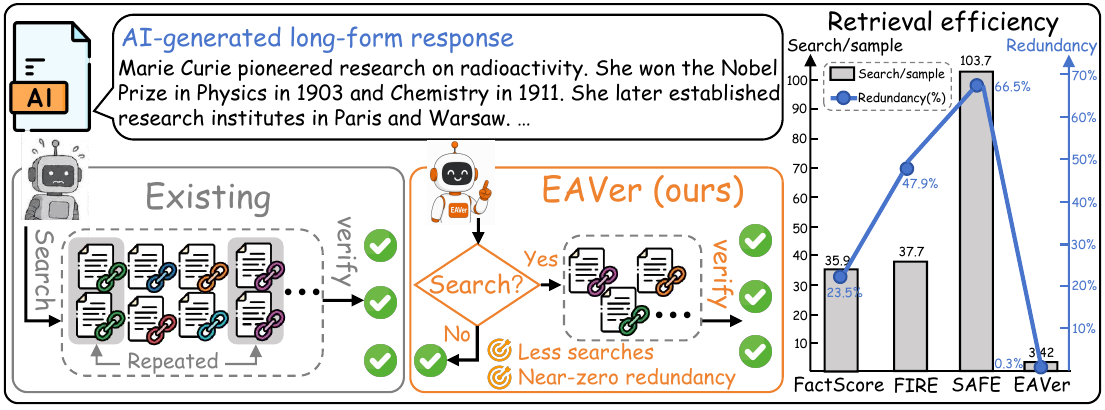}
\caption{\textbf{Search efficiency of \method.}
\emph{Left:} Existing per-claim evaluators repeatedly search for overlapping evidence; \method adaptively searches and reuses evidence across related claims.
\emph{Right:} Mean searches per sample and link redundancy on an independent VeriFastScore validation split. Link redundancy is the fraction of links repeated within the same evaluation sample.}
\label{fig:teaser}
\end{figure}

Existing long-form factuality evaluators generally follow a decompose-search-verify recipe: extract factual claims, search for external evidence, and predict whether each claim is supported~\citep{min2023factscore,wei2024longform,song2024veriscore}.
Subsequent work broadens the range of verifiable content, uses iterative search for difficult claims, or batches extraction and verification for greater efficiency~\citep{xie2025fire,rajendhran2025verifastscore}.
Nevertheless, search control and verification remain hand-orchestrated or separately optimized rather than jointly learned as a unified policy for response-level evaluation.

Despite their different implementations, existing methods share two structural limitations.
Most hard-code verification as an orchestration of separately prompted LLM and search-API calls, and process each claim largely in isolation.
Consequently, the number of calls grows with the number of claims, while related claims repeatedly search for overlapping evidence about the same entities and events.
To quantify this inefficiency, we evaluate each method on an independent VeriFastScore validation split and record searches per sample and link redundancy.
Figure~\ref{fig:teaser} illustrates this structural contrast and quantifies its cost: prior per-claim pipelines issue 35.9--103.7 searches per sample with 23.5--66.5\% repeated links, whereas \method uses 3.42 searches with only 0.3\% redundancy.
The high redundancy of prior pipelines persists even when exact query strings differ, showing that per-claim control and string-level caching do not fully address cross-claim evidence reuse.
More fundamentally, a fixed controller cannot be optimized as a response-level policy that learns when to search, what to search for, and when available evidence is already sufficient.
Although VeriFastScore trains its verifier, it consumes evidence collected by a separate search stage and therefore does not learn an interactive policy for search and evidence reuse~\citep{rajendhran2025verifastscore}.
Motivated by language-model agents that interleave reasoning with external actions~\citep{yao2023react,schick2023toolformer}, we introduce \method, an end-to-end agentic verifier that learns to control the complete response-level verification workflow.
Rather than executing a fixed sequence of modules, \method represents verification as a single trajectory that jointly organizes claims, acquires evidence, and produces factuality judgments.
It groups semantically related claims, routes each group to direct verification or targeted search based on model confidence, and compresses evidence returned by search into compact in-context memos that later claims can reuse directly from the shared interaction history.
Learning this structured behavior requires supervision beyond final factuality labels.
Building on synthetic supervision for instruction following~\citep{wang2023selfinstruct}, we develop a policy-aware synthesis pipeline in which a privileged teacher uses gold atomic claims and labels to generate executable multi-turn tool-interaction trajectories with live search rather than post-hoc explanations; the student receives none of these gold annotations at inference time.
We apply quality-control filters for structural validity, label alignment, proper tool use, search-budget compliance, and the absence of label leakage, retaining only demonstrations that faithfully instantiate the complete verification policy.

Base verifiers are strongly biased toward the supported class, making false support particularly consequential: it allows unsupported content to pass as factual.
We therefore first learn the core policy from executable verification demonstrations and study decision-focused Direct Preference Optimization (DPO)~\citep{rafailov2023dpo} only as a conservative extension that restricts preference learning to factuality decisions within otherwise matched trajectories, without directly optimizing the learned search and evidence-use behavior.
Experiments show that \method achieves the strongest Macro-F1 among search-based systems on both benchmarks while using about one fifth as many searches as the most search-efficient baseline. Moreover, the proposed policy-learning framework yields consistent gains across backbone generations, parameter scales, and model families.
Our contributions are:
\begin{itemize}
    \item[\ding{182}]\textbf{Empirical finding.} We systematically analyze the search efficiency of representative long-form factuality verifiers, uncovering substantial per-sample search overhead and cross-claim evidence redundancy.

    \item[\ding{183}]\textbf{Policy-learning contribution.} We develop a privileged-teacher synthesis and quality-control framework that converts claim supervision into executable verification behavior, together with a controlled preference-construction procedure that isolates factuality-decision errors.

    \item[\ding{184}]\textbf{Methodological contribution.} We propose \method, an end-to-end agentic long-form verifier that unifies claim grouping, adaptive search, in-context evidence reuse, and factuality prediction within a single trainable policy.

    \item[\ding{185}]\textbf{Empirical validation.} \method achieves the strongest Macro-F1 among search-based systems on both benchmarks while using about one fifth as many searches as the most search-efficient baseline, with consistent improvements across backbone generations, parameter scales, and model families.
\end{itemize}
\begin{figure*}[t]
\centering
\includegraphics[width=0.92\textwidth]{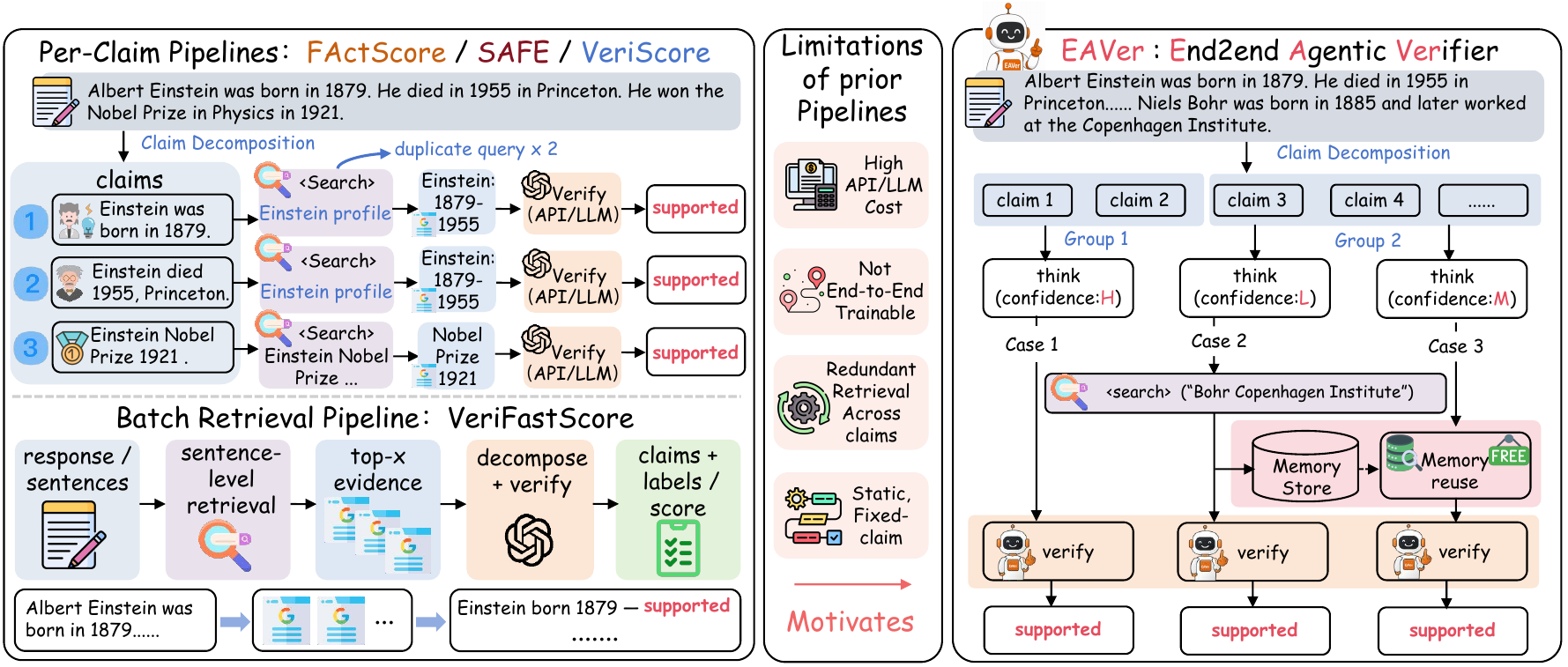}
\caption{\textbf{Fixed verification pipelines versus \method.}
\emph{Left:} FActScore, SAFE, and VeriScore verify claims independently; VeriFastScore batches sentence-level search.
\emph{Center:} Fixed pipelines have four limitations: high API and LLM costs, no end-to-end training, cross-claim search redundancy, and static claim processing.
\emph{Right:} \method groups related claims, routes them to direct verification or search by confidence, and stores evidence for reuse.}
\label{fig:architecture}
\end{figure*}

\section{Related Work}

\paragraph{Long-form factuality evaluation.}
FActScore established fine-grained evaluation by decomposing a generation into atomic facts and measuring the fraction supported by a reliable knowledge source~\citep{min2023factscore}.
FacTool generalized tool-augmented factuality detection across tasks through claim extraction, query generation, evidence collection, and agreement checking~\citep{chern2023factool}, while SAFE operationalized long-form evaluation with LLM-based decomposition and web-search verification~\citep{wei2024longform}.
VeriScore further distinguishes verifiable claims from subjective or otherwise uncheckable content~\citep{song2024veriscore}.
More recent systems improve either search adaptivity or throughput.
FIRE iteratively searches for and verifies evidence until an individual claim can be decided~\citep{xie2025fire}, whereas VeriFastScore uses sentence-level batch search and a fine-tuned model to jointly extract and verify claims~\citep{rajendhran2025verifastscore}.
However, their overall control remains claim-local or stage-wise, with search and verification either prompt-orchestrated or optimized separately.
Our work instead studies a joint policy for response-level search and verification.

\paragraph{Learned search agents.}
Recent work trains language models to treat search as part of the reasoning policy rather than a fixed preprocessing step.
R1-Searcher and Search-R1 use outcome-based reinforcement learning to elicit autonomous, multi-turn search~\citep{song2025r1searcher,jin2025searchr1}, while ReSearch learns interleaved reasoning and search without supervised reasoning traces~\citep{chen2025research}.
Smart-Searcher combines an SFT cold start with reinforcement learning to encourage dynamic use of internal and external knowledge~\citep{song2025smartsearcher}.
These methods primarily target knowledge-intensive question answering, where search supports a single final answer.
Long-form factuality verification instead requires exhaustive coverage of claims across an entire response, aligned claim-level judgments, and control of cumulative search cost; our work studies learned search under this multi-claim objective.

\section{Task Definition}
\label{sec:task}

The input is a question $q$ and a model response $y$.
For evaluation, the response is annotated with a gold set of factual claims
$C=\{c_i\}_{i=1}^{n}$, where each gold claim has a binary label
\[
z_i \in \{\support,\unsupport\}.
\]
The system must predict a claim set
$\widehat{C}=\{\widehat{c}_j\}_{j=1}^{\widehat{n}}$ and assign each predicted
claim a label $\widehat{z}_j \in \{\support,\unsupport\}$.
In our current evaluation, predicted claims are greedily matched one-to-one to
gold claims using string similarity with threshold 0.5.
For matched claims, predicted labels are scored against the corresponding gold labels.
We report claim-extraction F1, macro-F1, and False Support Rate (FSR).

\paragraph{Claim extraction F1.}
Let $M$ be the set of matched predicted--gold claim pairs.  We compute
extraction precision, recall, and their harmonic mean as
\[
P_{\mathrm{ext}}=\frac{|M|}{|\widehat{C}|},\qquad
R_{\mathrm{ext}}=\frac{|M|}{|C|},\qquad
F_{1}^{\mathrm{ext}}=
\frac{2P_{\mathrm{ext}}R_{\mathrm{ext}}}
{P_{\mathrm{ext}}+R_{\mathrm{ext}}}.
\]
The last quantity is reported as \texttt{claim\_ext\_F1} in
Table~\ref{tab:main_results}.
The combined metric evaluates decomposition independently of factuality labels
and penalizes both omitted gold claims and over-decomposition with unmatched
predictions.

\paragraph{Macro-F1.}
Macro-F1 is the average of F1 for \support and F1 for \unsupport.
It is the primary score because the supported class dominates the validation set.

\paragraph{False Support Rate (FSR).}
FSR is the fraction of gold \unsupport claims incorrectly labeled \support:
\[
\mathrm{FSR} = \frac{\#(\mathrm{gold}=\unsupport,\ \mathrm{pred}=\support)}{\#(\mathrm{gold}=\unsupport)}.
\]
Lower FSR is better because false support allows unsupported content to be accepted as factual.
FSR therefore isolates the asymmetric error that is most consequential for a factuality verifier.
Macro-F1 summarizes performance across both classes, so reporting the two metrics together captures both balanced class performance and false-support risk.

\section{\method}
\label{sec:method}

\method is designed around three principles:
(1) claim verification should be optimized as a single trajectory rather than a hand-written pipeline,
(2) search should be conditional on model confidence, and
(3) evidence should be reusable across claims that share entities or events.

\subsection{Policy-Aware Verification Trajectory Synthesis}

Final-label supervision offers little guidance for the behavior preceding a factuality decision.
For example, the zero-shot Qwen3-8B verifier in Table~\ref{tab:main_results} has FSRs of 89.18\% on VeriFastScore and 89.72\% on FaStFact-Bench, demonstrating a consistent tendency to accept unsupported claims as factual before policy training.
Search access alone does not teach the model when to doubt a plausible claim, seek counter-evidence, or overturn the response.
Rather than distilling only final labels, we synthesize executable demonstrations that jointly supervise claim decomposition, confidence-conditioned search, query formulation, evidence summarization, and negative verification.

\begin{figure*}[!t]
\centering
\includegraphics[width=0.90\textwidth]{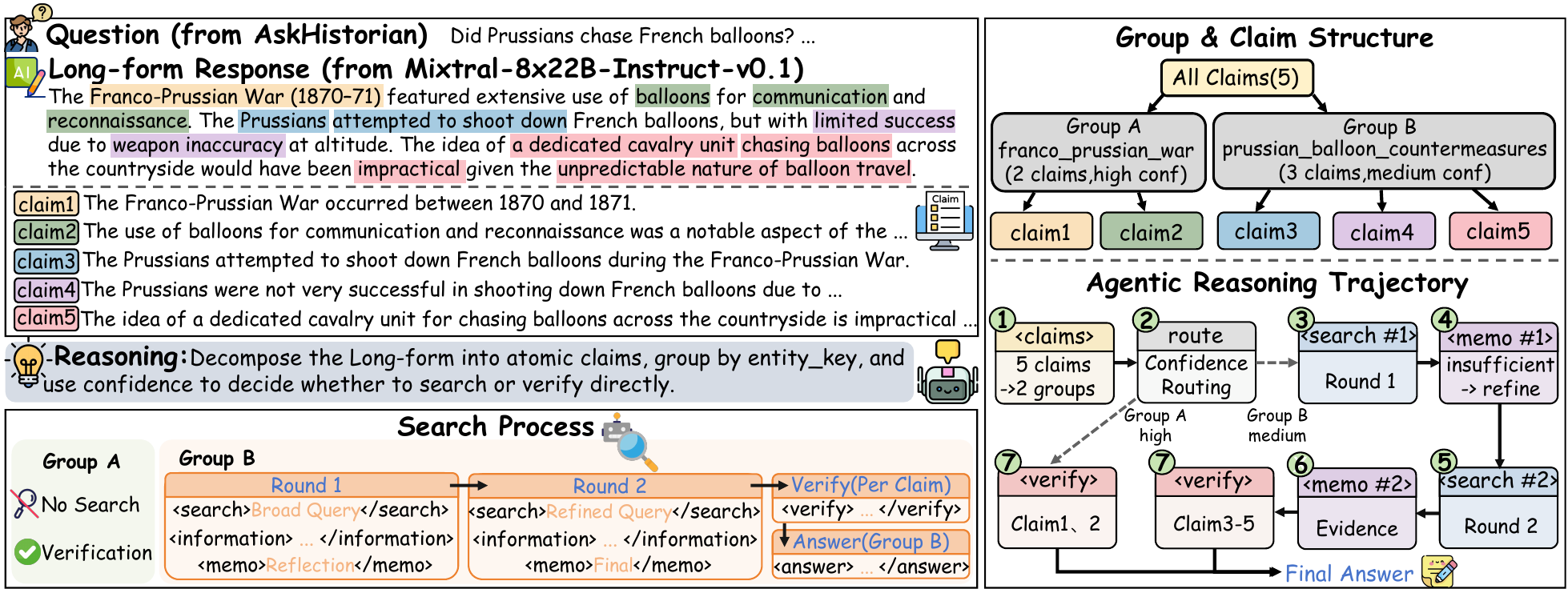}
\caption{\textbf{A representative \method inference trajectory.}
Five claims form two entity groups. The high-confidence group is verified directly; the medium-confidence group uses two searches and shared evidence before all five verdicts are returned.}
\label{fig:inference_trajectory}
\end{figure*}

Figure~\ref{fig:inference_trajectory} illustrates one such real trajectory from the in-domain VeriFastScore dataset.

We use Claude Opus 4.8 as a privileged teacher during data construction.
It privately receives gold atomic claims and labels to construct label-consistent trajectories; the student sees neither at inference.
The teacher copies each claim verbatim into a \texttt{<claims>} JSON array and assigns two policy fields: \texttt{confidence} (\emph{high}, \emph{medium}, or \emph{low}) and an \texttt{entity\_key}.
Claims sharing an entity key are processed together.
An all-high-confidence group is verified directly in a batch, whereas any medium- or low-confidence claim triggers one focused search for the group.
The trajectory uses five structured tags: \texttt{<claims>}, \searchtag, \memotag, \verifytag, and \answertag.
When the teacher ends a turn with \texttt{<search>query</search>}, generation pauses; the runtime parses the query, calls the search tool, injects its results as an \texttt{<information>} environment message, and resumes with full history.
This yields multi-turn tool-interaction transcripts rather than post-hoc textual rationales alone.

After search, the teacher compresses the long, noisy observation into a compact \memotag{} evidence summary before emitting one \verifytag{} decision per claim and an aligned \answertag{} JSON.
The memo is an explicit evidence-summary scaffold, not an external cache or lookup store: it remains in the shared dialogue context so that claims in the same entity group can implicitly reuse the evidence without a separate recall action.
We retain trajectories that pass structural, label-alignment, tool-use, search-budget, and private-key leakage checks.
Twenty human experts additionally conducted random spot checks of the retained trajectories, with 98\% of inspected trajectories passing the audit.
The retained demonstrations provide process-level supervision for claim organization, evidence acquisition and reuse, and aligned factuality decisions.

\begin{table}[h]
\centering
\footnotesize
\setlength{\tabcolsep}{2.5pt}
\renewcommand{\arraystretch}{1.08}
\begin{tabular*}{\columnwidth}{@{\extracolsep{\fill}}lrrrrr@{}}
\toprule
Stage
& \shortstack{Items}
& \shortstack{Searches\\/traj.}
& \shortstack{Length\\(k chars)}
& Sup.
& Unsup. \\
\midrule
SFT trajectories   & 1{,}447 & 2.24 & 9.16 & 70.4\% & 29.6\% \\
DPO pairs (chosen) & 794     & 3.69 & 13.80 & 61.7\% & 38.3\% \\
\bottomrule
\end{tabular*}
\caption{\textbf{Synthetic training-data overview.}
Trajectory length counts generated text and injected search-result observations
after the initial prompt, giving a tokenizer-independent measure.
Label shares are computed over 19{,}334 SFT decisions and 15{,}686 decisions
in the DPO chosen completions.
Each rejected DPO completion shares the chosen completion's searches and
evidence and differs only at corrected decision labels; the 794 pairs comprise
405 false-support-to-unsupported and 389 false-unsupported-to-supported pairs, respectively.}
\label{tab:training_data_overview}
\end{table}

\begin{table*}[t]
\centering
\small
\setlength{\tabcolsep}{4.5pt}
\begin{tabular}{lc|ccc|ccc}
\toprule
\multirow{2}{*}{\textbf{Method}} & \multirow{2}{*}{\textbf{Searches/sample}}
& \multicolumn{3}{c|}{\textbf{VeriFastScore (\%)}}
& \multicolumn{3}{c}{\textbf{FaStFact-Bench (\%)}} \\
\cmidrule(lr){3-5} \cmidrule(lr){6-8}
 &  & Macro F1 & FSR $\downarrow$ & claim\_ext\_F1 & Macro F1 & FSR $\downarrow$ & claim\_ext\_F1 \\
\midrule
\rowcolor{tablegroupgray}
\multicolumn{8}{c}{\textbf{Zero-shot LLM verifiers}} \\ \midrule
Qwen2.5-7B-Instruct                 & --        & 43.61 & 96.71 & 64.69 & 48.58 & 96.66 & 71.30 \\
Qwen3-8B                   & --        & 49.54 & 89.18 & 65.63 & 54.30 & 89.72 & 71.83 \\
Qwen3-max                  & --        & 60.07 & 74.65 & 73.75 & 70.93 & 58.64 & 70.12 \\
GPT-4.1                    & --        & 61.30 & 74.23 & 69.14 & 70.73 & 61.45 & 72.11 \\
Claude Sonnet 4.6          & --        & 62.66 & 72.53 & 74.17 & 70.75 & 41.18 & 62.95 \\
Claude Opus 4.8            & --        & 65.11 & 68.79 & 73.37 & 72.21 & 42.11 & 65.28 \\
\midrule
\rowcolor{tablegroupgray}
\multicolumn{8}{c}{\textbf{Search-augmented pipelines and task-trained verifiers}} \\ \midrule
FActScore~\citep{min2023factscore}        & 35.88 & 64.83 & 56.57 & 59.43 & 64.58 & 47.39 & 30.97 \\
SAFE~\citep{wei2024longform}              & 106.90 & 63.40 & 64.01 & 58.92 & 62.10 & 65.60 & 56.82 \\
VeriScore~\citep{song2024veriscore}       & 19.18 & 63.38 & 55.47 & 77.53 & 60.89 & 50.25 & 46.58 \\
FIRE~\citep{xie2025fire}                  & 37.69 & 63.40 & 64.00 & 58.92 & 69.32 & 48.13 & 32.04 \\
VeriFastScore~\citep{rajendhran2025verifastscore} & 15.62 & 70.55 & \textbf{38.28} & 81.25 & 64.99 & 36.32 & 46.10 \\
\midrule
\rowcolor{tablegroupgray}
\multicolumn{8}{c}{\textbf{\emph{\textbf{E}nd-to-end \textbf{A}gentic \textbf{Ver}ifier}}} \\
\method (Qwen2.5-7B-Instruct)            & 4.21 & 71.77 & 38.51 & 80.52 & 66.01 & 41.20 & 71.82 \\
\method (Llama-3.1-8B)                   & 3.78 & 72.40 & 40.90 & 80.77 & 66.54 & 58.53 & 73.39 \\
\method (Qwen3-8B)            & \textbf{3.14} & \textbf{73.43} & 41.57 & \textbf{81.48} & \textbf{74.05} & \textbf{35.87} & \textbf{74.24} \\
\bottomrule
\end{tabular}
\caption{\textbf{Main results.}
Results on \textbf{VeriFastScore} (in-domain) and \textbf{FaStFact-Bench} (out-of-domain, human-annotated).
Searches/sample averages search-API requests over both datasets; LLM calls are excluded, and ``--'' denotes no external search.
Evaluation metrics are percentages. Best reported values in each column are \textbf{bold}.}
\label{tab:main_results}
\end{table*}

\subsection{Decision-Focused Preference Post-training}

After supervised training on our policy-aware synthetic trajectories, the model has learned the complete verification policy, including claim organization, search, evidence use, and factuality prediction.
Preference post-training is therefore used only as an optional correction for residual verdict errors.
Applying trajectory-level preference optimization to independently sampled completions would entangle the desired label correction with incidental differences in search and evidence-use behavior.
Among models with comparable \fscore, we prefer the one with lower \fsr: false support allows unsupported content to pass as factual, and identifying such negative cases is therefore more consequential for a verifier than confirming additional supported ones.
We use preference post-training to seek an operating point with lower \fsr while constraining changes to overall \fscore, and localize the update through both controlled preference construction and a decision-focused loss.

Specifically, we analyze rollouts produced by the policy trained on the synthetic trajectory corpus and select trajectories with incorrect verdicts for which the collected evidence is sufficient to support the gold decision.
In these cases, the model has already obtained the evidence needed for verification but still emits an incorrect label, indicating a decision failure rather than a search failure.
Rather than regenerate the full reasoning trajectory, we keep the trajectory fixed and correct only the erroneous verdict.
For each selected case, the original erroneous completion serves as the rejected response, while the chosen response is created by correcting the affected \verifytag{} labels and synchronized \answertag{} entries to the gold decisions.
Claim grouping, search queries, search observations, evidence summaries, and all non-decision text remain identical.
The resulting corpus contains 794 pairs spanning both false-support-to-unsupported and false-unsupported-to-supported corrections.
Because the surrounding completion is matched, the preference signal isolates the factuality decision rather than differences in tool use or language realization.

We further restrict the DPO log-probability calculation to the decision tokens expressing \support or \unsupport inside \verifytag{} and \answertag.
Let $m_t$ be a binary mask that is one only for these decision tokens.
The resulting masked sequence score is
\begin{equation}
s_\theta^M(y\mid x)
=
\sum_{t=1}^{|y|}m_t
\log \pi_\theta(y_t\mid x,y_{<t}).
\label{eq:masked_sequence_score}
\end{equation}
Defining the reference-relative score as
$r_\theta^M(y,x)=s_\theta^M(y\mid x)-s_{\mathrm{ref}}^M(y\mid x)$,
and the preference margin as
$\Delta r_\theta^M=r_\theta^M(y^+,x)-r_\theta^M(y^-,x)$,
our decision-focused objective is
\begin{equation}
\mathcal{L}_{\mathrm{DF\text{-}DPO}}
=
-\mathbb{E}_{(x,y^+,y^-)}
\log \sigma\!\left(
\beta\Delta r_\theta^M
\right).
\label{eq:decision_focused_dpo}
\end{equation}
All claim text, search, evidence, memo, and formatting tokens therefore have
$m_t=0$ and are excluded from the preference loss; conventional DPO is
recovered by setting $m_t=1$ for every completion token.
Spanning both correction directions prevents the training signal from reducing \fsr simply by pushing every claim toward \unsupport.
This optional stage directly supervises only the verdict tokens and does not reward changes to the preceding search trajectory.
Section~\ref{sec:results} evaluates whether this localized update reduces false support without degrading balanced factuality performance.

\section{Experimental Setup}
\label{sec:setup}

\paragraph{Backbone.}
We conduct our main evaluation of \method on Qwen2.5-7B-Instruct~\citep{qwen2025qwen25} and Qwen3-8B~\citep{yang2025qwen3}, providing comprehensive validation across two strong open-weight models.
Additional ablations span model scales from 4B to 32B and multiple model families, including Llama 3.1~\citep{grattafiori2024llama3}, to assess the generalizability of our synthesized trajectories.

\paragraph{Search setting.}
Following Search-R1~\citep{jin2025searchr1}, we implement the search actions used during training with a local dense-search backend, enabling stable and reproducible interaction trajectories.
Queries are encoded with E5-base-v2~\citep{wang2022e5}, and a FAISS index~\citep{johnson2017faiss} returns the top three search results.
This fixed local search backend avoids fluctuations in the latency, availability, and outputs of external search APIs.
For evaluation, we keep the model checkpoint, \searchtag{} action protocol, and decoding configuration fixed, but replace the local search backend with the Google Search API without retraining the model.

\paragraph{Data.}
We evaluate on two benchmarks.
For in-domain evaluation, we construct a clean test set from the official 5,900-example test split of VeriFastScore~\citep{rajendhran2025verifastscore}.
We remove persona and creative-writing sources, abstentions and invalid responses, examples without valid binary claims, and any sample whose question--response fingerprint matches data used for training, trajectory synthesis, rollout generation, or development.
The resulting set contains 22,532 claims, with zero sample-level fingerprint overlap with all touched pools.
For out-of-distribution evaluation, we use the independently released, human-annotated FaStFact-Bench~\citep{wan2025fastfact}.
We map its fine-grained labels to our binary scheme and discard non-factual housekeeping annotations, abstentions, and samples without verifiable claims, yielding 6,954 claims.

\paragraph{Baselines.}
We compare against two baseline families shown in Table~\ref{tab:main_results}.
The search-free group, which uses no external search, includes LLM-only Qwen2.5-7B-Instruct~\citep{qwen2025qwen25}, LLM-only Qwen3-8B~\citep{yang2025qwen3}, Qwen3-Max, GPT-4.1, Claude Sonnet 4.6, and Claude Opus 4.8, each using one model invocation per sample.
The search-augmented group comprises FActScore~\citep{min2023factscore}, SAFE~\citep{wei2024longform}, VeriScore~\citep{song2024veriscore}, FIRE~\citep{xie2025fire}, and VeriFastScore~\citep{rajendhran2025verifastscore}.
All experiments are run three times, and we report the mean over the three runs.

\section{Main Results}
\label{sec:results}

Table~\ref{tab:main_results} reports the main comparison on two held-out benchmarks under a unified evaluation protocol.
We group methods by category: zero-shot LLM verifiers without search, search-augmented pipelines with GPT-4.1 backends, and \method.

\paragraph{Effectiveness.}
We select Qwen2.5-7B-Instruct, Qwen3-8B, and Llama-3.1-8B as three representative backbones from the Qwen and Llama families to evaluate the effectiveness of \method across model architectures.
Among them, \method (Qwen3-8B) achieves the best overall Macro-F1 on both the in-domain VeriFastScore benchmark and the out-of-domain FaStFact-Bench, reaching 73.43 and 74.05, respectively.
Compared with the search-free Qwen3-8B verifier, training on our agentic verification trajectories reduces FSR from 89.18 to 41.57 on VeriFastScore and from 89.72 to 35.87 on FaStFact-Bench.
This pronounced reduction shows that the gain is not inherited from the backbone alone: the trained policy learns to use evidence to identify unsupported claims rather than defaulting to \support.
\method (Qwen3-8B) also obtains the highest claim extraction F1 on both benchmarks.
By contrast, most prior search-augmented pipelines exhibit a substantial drop in claim extraction F1 out of domain.
Our case inspection suggests that their static, LLM-driven decomposition stages tend to over-extract fine-grained or redundant claims, inflating $|\widehat{C}|$ without a commensurate increase in matched gold claims and thereby lowering extraction precision and F1.
VeriFastScore attains the lowest in-domain FSR, but this result should be interpreted in light of its two-stage fine-tuning on roughly 9K synthetic prompt--response pairs and its collection of sentence-level evidence before claim decomposition~\citep{rajendhran2025verifastscore}.
Finally, Claude Opus 4.8 achieves a strong 72.21 Macro-F1 on FaStFact-Bench as a search-free verifier, providing empirical support for its use as our privileged teacher.
Although neither trajectory synthesis nor post-training uses FaStFact-Bench examples or annotations, the resulting Qwen3-8B student reaches 74.05 Macro-F1, suggesting that the teacher-guided agentic supervision transfers beyond its source data.

\paragraph{Cost-efficient verification.}
For a like-for-like comparison, Table~\ref{tab:main_results} counts external search requests only.
\method (Qwen3-8B) averages 3.14 searches per sample, about 80\% fewer than the 15.62 used by VeriFastScore, the most search-efficient baseline.
This measure does not include the LLM calls incurred by multi-stage orchestration.
For example, SAFE can require up to $2 + 5 \times (1+1) + 1 = 13$ calls for a single relevant claim: 2 LLM calls for self-containment and relevance, 5 query-generation calls paired with 5 Google Search calls, and 1 final verdict call~\citep{wei2024longform}.
In contrast, \method groups related claims, directly verifies high-confidence groups, searches only for uncertain groups, and reuses evidence returned by search through in-context memos, allowing one search to support decisions for several related claims at once.

\begin{figure}[h]
\centering
\includegraphics[width=\columnwidth]{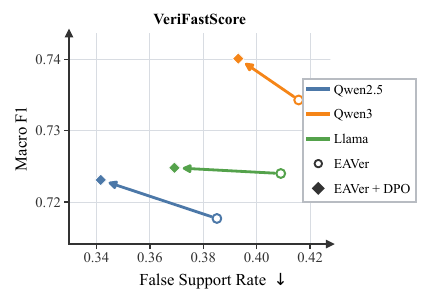}
\caption{\textbf{Decision-focused DPO lowers FSR for \method on VeriFastScore.}
Results are reported on VeriFastScore across three backbones.
Circles show \method before DPO, and diamonds show \method after DPO.
Arrows connect matched checkpoints; leftward movement indicates lower FSR, while upward movement indicates higher Macro-F1. Colors identify backbones.}
\label{fig:dpo_pareto}
\end{figure}

\begin{figure*}[!t]
\centering
\includegraphics[width=\textwidth]{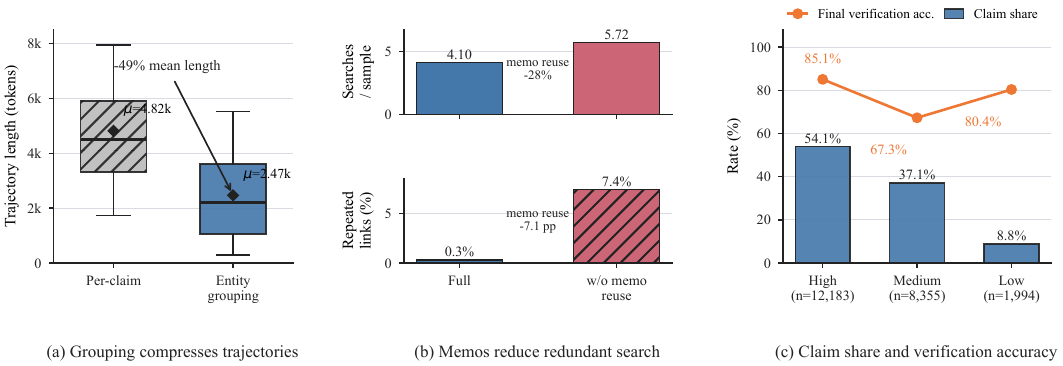}
\caption{\textbf{Mechanism-level component analysis for Qwen3-8B.}
(a) On FaStFact-Bench, entity grouping reduces token-level trajectory length per sample.
(b) On FaStFact-Bench, disabling memo-based evidence reuse increases searches per sample and repeated-link redundancy.
(c) On VeriFastScore, bars show claim share and final verification accuracy by confidence level.}
\label{fig:ablation_bars}
\end{figure*}

\paragraph{Decision-focused preference post-training.}
Figure~\ref{fig:dpo_pareto} evaluates decision-focused DPO applied to \method on VeriFastScore across three backbones under fixed search and decoding settings.
Across all three backbones, applying DPO to \method lowers FSR while preserving or modestly improving Macro-F1.
For Qwen2.5-7B, which exhibits the largest FSR reduction, FSR drops from 38.51 to 34.16, while Macro-F1 rises from 71.77 to 72.31.
The consistent upper-left shifts show that preference post-training strengthens the ability of \method to reject unsupported claims rather than merely shifting the prediction prior.

\section{Analysis and Extensions}
\label{sec:analysis}

We organize the remaining experiments around two questions: whether \method's core components jointly enable efficient and reliable verification, and whether the learned agentic policy generalizes across model generations, parameter scales, and families alike.

\subsection{Component Ablations}

\begin{table}[h]
\centering
\footnotesize
\setlength{\tabcolsep}{2.2pt}
\begin{tabular}{@{}lcccc@{}}
\toprule
\textbf{Policy variant}
& \shortstack{\textbf{Macro F1}\\(\%) $\uparrow$}
& \shortstack{\textbf{FSR}\\(\%) $\downarrow$}
& \shortstack{\textbf{Avg. traj.}\\\textbf{length} $\downarrow$}
    & \shortstack{\textbf{Searches}\\\textbf{/ sample} $\downarrow$} \\
\midrule
\textbf{Full policy}
& \textbf{74.05} & \textbf{35.87} & \textbf{2,474} & \textbf{4.10} \\
\midrule
\quad w/o entity grouping
& 71.82 & 39.41 & 4,821 & 6.35 \\
\quad w/o confidence gate
& 72.19 & 36.07 & 4,058 & 7.65 \\
\quad w/o memo reuse
& 72.41 & 40.29 & 4,260 & 5.72 \\
\bottomrule
\end{tabular}
\caption{\textbf{Component ablations.}
Entity grouping shortens trajectories, confidence routing avoids unnecessary search, and memo reuse limits redundant searches.}
\label{tab:component_ablation}
\end{table}

Figure~\ref{fig:ablation_bars}(a,b) and Table~\ref{tab:component_ablation} report FaStFact-Bench ablations, while panel (c) diagnoses confidence behavior on VeriFastScore.
In panel (a), grouping cuts trajectory length from 4,821 to 2,474 tokens (49\%) through shared claim context.
In panel (b), disabling memos increases searches per sample from 4.10 to 5.72 and repeated links from 0.3\% to 7.4\%, confirming that evidence reuse avoids redundant retrieval.
Panel (c) shows a non-monotonic relation between confidence and accuracy: high-, medium-, and low-confidence claims reach 85.1\%, 67.3\%, and 80.4\%, respectively.
Medium-confidence claims can be deceptively familiar, prompting broad yet non-discriminative queries.
For example, verifying ``Einstein won the Nobel Prize in 1922'' may search pages that mention both the 1921 prize and its 1922 presentation, making the award year easy to misread.
Low-confidence claims often contain niche names or technical terms that support narrower queries and better-matched evidence.

\subsection{Generalization of Model Families and Sizes}

Using the same training and evaluation configuration, we test \method on six Qwen and Llama models ranging from 4B to 32B on the out-of-domain FaStFact-Bench.

\begin{figure}[h]
\centering
\includegraphics[
  width=0.82\columnwidth
]{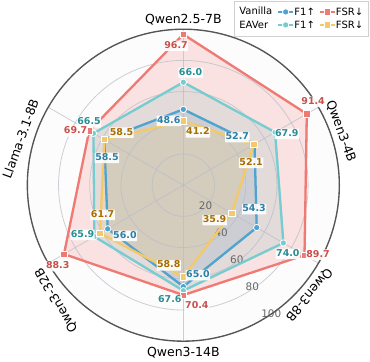}
\caption{\textbf{Cross-backbone generalization in Macro-F1 and FSR.}
Base and \method-trained models are evaluated on the out-of-distribution FaStFact-Bench across six Qwen and Llama backbones.
Higher Macro-F1 and lower FSR indicate better performance; \method improves both metrics for every tested model and scale.}
\label{fig:backbone_ood_transfer}
\end{figure}

Figure~\ref{fig:backbone_ood_transfer} shows that \method consistently improves long-form verification across both model families and at every tested scale, increasing Macro-F1 and reducing FSR for all six backbones.
The improvement magnitude varies across backbones, while EAVer improves Macro-F1 and reduces FSR for all six models.
Qwen2.5-7B, the weakest base model in our evaluation, achieves a 17.4-point Macro-F1 improvement after policy training.
This demonstrates policy transfer across model generations, families, and scales without backbone-specific synthesis.

\section{Conclusion}

Privileged-teacher trajectories train \method to jointly learn claim grouping, selective search, and evidence reuse, improving factuality with about 80\% fewer searches and transferring across model families and scales.
This supports jointly learning search, evidence retention, and factuality decisions.

\clearpage
\bibliography{references}

\clearpage
\appendix
\section{Appendix: Evaluation and Reproducibility Details}
\label{sec:appendix-evaluation}

\paragraph{Evaluation protocol.}
The agent generates a complete trajectory containing \texttt{<claims>}, \searchtag, \memotag, and \verifytag blocks, and concludes with an \answertag JSON listing all predicted claims and labels.
Predicted claims are matched to gold claims by string similarity (\texttt{SequenceMatcher}, threshold 0.5); matched claims contribute to confusion-matrix metrics.
We report \emph{claim coverage} (fraction of gold claims matched), \emph{Macro F1} (mean of supported-F1 and unsupported-F1, our primary metric), \emph{FSR} (the fraction of gold-\unsupport{} claims that are predicted as \support), \emph{searches per sample}, and \emph{searches per claim} (SPC).
Agentic evaluation uses \texttt{max\_turns}$=20$ and \texttt{max\_new\_tokens}$=8192$.

\section{Training Design and Hyperparameters}
\label{sec:appendix-training}

This section provides the optimization, tokenization, and systems details for
the two policy-training stages.  The first stage teaches the complete
verification behavior from executable demonstrations.  The second stage starts
from the resulting policy and applies a conservative preference update to
factuality decisions.  All backbone comparisons use the same 1{,}447
demonstrations, action schema, and optimization settings unless an
architecture-specific memory constraint is stated explicitly.

\subsection{Policy Demonstration Training}
\label{sec:appendix-policy-training}

\paragraph{Training objective.}
Each training item is a complete multi-turn interaction consisting of a system
instruction, a question and long-form model response, assistant actions,
retrieval observations, evidence memos, claim-level decisions, and the final
structured answer.
We optimize causal language-model likelihood only on assistant outputs.
Tokens belonging to the system, user, and retrieval-environment turns are
assigned the ignore index, as is the assistant role prefix.
The supervised target contains the assistant content and its end-of-turn
marker.  Consequently, the policy learns to generate \texttt{<claims>},
\searchtag, \memotag, \verifytag, and \answertag{} outputs while conditioning
on, but not imitating, the question, source response, or tool observation.
This assistant-only objective also prevents long retrieved passages from
dominating the loss.

\paragraph{Sequence construction.}
We serialize each sample with the native conversation markers of its backbone
family and preserve all turns in chronological order.
For Qwen2.5 and Qwen3, each message is enclosed by
\texttt{<|im\_start|>} and \texttt{<|im\_end|>} markers.
For Llama~3.1, we use the corresponding header and end-of-turn tokens and omit
the auxiliary knowledge-date preamble so that training and evaluation use
identical templates.
We do not insert a separate reasoning marker and do not pack multiple samples
into one sequence.
Sequences are right-truncated to 8{,}192 tokens and padded to a multiple of
eight; if a tokenizer has no dedicated padding token, its end-of-sequence token
is used for padding.

\begin{table*}[t]
\centering
\footnotesize
\setlength{\tabcolsep}{5.0pt}
\renewcommand{\arraystretch}{1.08}
\begin{tabular}{lll}
\toprule
\textbf{Category} & \textbf{Setting} & \textbf{Value} \\
\midrule
Data
& Demonstrations
& 1{,}447 multi-turn policy trajectories \\
Sequence
& Maximum sequence length
& 8{,}192 tokens, no sample packing \\
Objective
& Token supervision
& Assistant content and assistant end-of-turn marker only \\
Optimization
& Optimizer
& AdamW \\
Optimization
& Learning rate and schedule
& $2\times10^{-5}$, cosine decay \\
Optimization
& Warmup and weight decay
& 0.05 warmup ratio, 0.01 weight decay \\
Optimization
& Gradient clipping
& 1.0 \\
Batching
& Per-device batch size
& 1 \\
Batching
& Effective global batch
& 16 for models up to 14B; 8 for the 32B model \\
Numerics
& Precision and attention
& bfloat16, FlashAttention-2 \\
Memory
& Activation memory
& Gradient checkpointing, non-reentrant implementation \\
Distributed
& Data parallelism
& One node with eight GPUs \\
Checkpointing
& Save policy
& Once per epoch, model weights only, at most three checkpoints \\
Reproducibility
& Training seeds
& 42, 123, and 2027 \\
\bottomrule
\end{tabular}
\caption{\textbf{Shared policy demonstration training configuration.}
The same semantic training corpus and objective are used for all backbones.
Only the memory-sensitive settings listed in
Table~\ref{tab:appendix-backbone-training} vary.}
\label{tab:appendix-sft-hparams}
\end{table*}

\paragraph{Optimization and precision.}
Table~\ref{tab:appendix-sft-hparams} summarizes the shared configuration.
Training uses the TRL supervised fine-tuning trainer, bfloat16 model and
optimizer computation, FlashAttention-2, and gradient checkpointing with
non-reentrant recomputation.
We use AdamW with a peak learning rate of $2\times10^{-5}$, cosine decay,
a 0.05 warmup ratio, weight decay 0.01, and gradient clipping at 1.0.
The per-device batch size is one.
On eight GPUs, two gradient-accumulation steps give an effective global batch
of 16 for the models up to 14B parameters.
Each configuration is trained independently with seeds 42, 123, and 2027.
Logging is performed every five optimizer steps.
Checkpoints are written at epoch boundaries and contain model weights only,
which avoids retaining optimizer states solely for evaluation.

\paragraph{Distributed execution.}
All reported policies are trained on a single node with eight NVIDIA H20 GPUs,
each with approximately 95--96 GB of memory.
Models up to 14B parameters use DeepSpeed ZeRO-2 with optimizer-state CPU
offload.
The dense 32B model uses ZeRO-3, which shards parameters across all eight GPUs
and offloads optimizer states to host memory.
For the 32B setting, we additionally reduce gradient accumulation from two to
one and disable data-loader worker processes to keep peak host memory below the
node limit.
These changes affect systems memory only; the demonstration corpus, loss mask,
learning rate, schedule, and sequence length remain unchanged.

\begin{table*}[t]
\centering
\footnotesize
\setlength{\tabcolsep}{3.8pt}
\renewcommand{\arraystretch}{1.08}
\begin{tabular}{lcccccc}
\toprule
\textbf{Backbone}
& \textbf{Epochs run}
& \textbf{Selected}
& \textbf{Grad.\ accum.}
& \textbf{Global batch}
& \textbf{ZeRO stage}
& \textbf{Loader workers} \\
\midrule
Qwen2.5-7B-Instruct & 3 & 2 & 2 & 16 & 2 & 2 \\
Qwen3-4B            & 3 & 2 & 2 & 16 & 2 & 2 \\
Qwen3-8B            & 3 & 2 & 2 & 16 & 2 & 2 \\
Qwen3-14B           & 3 & 2 & 2 & 16 & 2 & 2 \\
Qwen3-32B           & 2 & 2 & 1 & 8  & 3 & 0 \\
Llama-3.1-8B-Instruct & 3 & 2 & 2 & 16 & 2 & 2 \\
\bottomrule
\end{tabular}
\caption{\textbf{Backbone-specific training settings.}
The selected epoch is used for the reported evaluations.
The 32B changes are imposed by host and device memory constraints rather than
backbone-specific tuning.}
\label{tab:appendix-backbone-training}
\end{table*}

\subsection{Decision-Focused Preference Training}
\label{sec:appendix-dpo-training}

The input is the 794-pair bidirectional corpus described in
the preference-pair construction below.
Each pair contains a chosen and rejected completion with identical claim
decomposition, tool interaction, retrieved evidence, memo content, and
formatting.
Only synchronized factuality labels differ.

The model attends to the full prompt and completion, including all search
observations, but the DPO sequence score is accumulated only over the
\support{} and \unsupport{} decision tokens inside \verifytag{} and
\answertag.
The decision-token mask therefore changes which positions contribute to the
preference objective without deleting contextual evidence from the forward
pass.
This distinction is essential: the policy can use the complete trajectory to
evaluate a label while the preference gradient cannot reward incidental
changes to search queries, evidence wording, or memo style.

\begin{table}[t]
\centering
\footnotesize
\setlength{\tabcolsep}{4.0pt}
\renewcommand{\arraystretch}{1.08}
\begin{tabular}{ll}
\toprule
\textbf{Setting} & \textbf{Value} \\
\midrule
Initialization & Selected policy checkpoint \\
Preference pairs & 794 \\
Epochs & 3 \\
Per-device batch size & 1 \\
Effective global batch & 32 \\
Learning rate & $5\times10^{-7}$ \\
Schedule & Cosine decay \\
Warmup ratio & 0.10 \\
Weight decay & 0 \\
DPO temperature $\beta$ & 0.1 \\
Loss type & Sigmoid DPO \\
Maximum total length & 5{,}120 tokens \\
Maximum prompt length & 2{,}048 tokens \\
Precision & bfloat16 \\
Optimizer & Paged 8-bit AdamW \\
Gradient checkpointing & Enabled, non-reentrant \\
Reference log probabilities & Precomputed \\
Auxiliary NLL coefficient & 0 \\
Random seed & 42 \\
\bottomrule
\end{tabular}
\caption{\textbf{Decision-focused preference training configuration.}}
\label{tab:appendix-dpo-hparams}
\end{table}

Table~\ref{tab:appendix-dpo-hparams} gives the complete preference-training
configuration.
We train for three epochs with a per-device batch of one and effective global
batch of 32 across eight GPUs.
The learning rate is $5\times10^{-7}$ with cosine decay and a 0.10 warmup ratio.
We use sigmoid DPO with $\beta=0.1$, zero weight decay, no auxiliary
negative-log-likelihood term, bfloat16 precision, gradient checkpointing, and a
paged 8-bit AdamW optimizer.
The maximum combined sequence length is 5{,}120 tokens and the maximum prompt
length is 2{,}048 tokens.
Reference-model log probabilities are precomputed with the same decision mask.
Models are saved once per epoch; final checkpoint selection uses the disjoint
development set and the constrained criterion described in the main paper.

\section{Policy Demonstration and Preference Data Construction}
\label{sec:appendix-data-construction}

This section describes the two training corpora at the level of their
statistical construction rather than implementation details.
We refer to the first corpus as \emph{policy demonstrations} because it
supervises the complete behavior of the verifier, not only its final
factuality labels.
The second corpus contains controlled \emph{preference pairs} that refine the
decision rule without changing the learned search policy.

\definecolor{promptbackground}{RGB}{246,249,252}
\definecolor{promptframe}{RGB}{108,126,145}
\definecolor{promptkeyword}{RGB}{18,79,123}
\lstdefinestyle{policy-prompt}{
  basicstyle=\ttfamily\tiny,
  backgroundcolor=\color{promptbackground},
  frame=single,
  rulecolor=\color{promptframe},
  breaklines=true,
  breakatwhitespace=false,
  columns=fullflexible,
  keepspaces=true,
  showstringspaces=false,
  captionpos=b,
  aboveskip=5pt,
  belowskip=5pt,
  xleftmargin=2pt,
  xrightmargin=2pt,
  framexleftmargin=3pt,
  framexrightmargin=3pt
}

\subsection{Policy Demonstration Synthesis}
\label{sec:appendix-demonstration-data}

\paragraph{Source pool and sampling.}
The source examples are drawn from the VeriFastScore training pool and contain
a question, a long-form model response, and an ordered list of atomic claims
with binary gold labels.
We retain only examples with at least one valid binary claim and no more than
40 claims.
Invalid responses, abstentions, and examples dominated by persona or creative
writing are excluded.
Sampling covers the eight source categories that occur in the corresponding
development distribution: ELI5, AskHistorian, FreshQA, FActScore, new books,
LongFacts, ShareGPT, and WritingPrompts.
Source quotas follow the development distribution rather than oversampling
examples with unsupported claims.
This preserves a realistic mixture of supported, unsupported, and mixed-label
samples.

Before synthesis, normalized question--response fingerprints are compared
against every development and test pool.
Any overlap is removed.
The same fingerprint audit is repeated after merging synthesis batches and
before model fitting.
Because the claim annotations are used as privileged synthesis inputs, they
are never used to select examples from the held-out evaluation benchmarks.

\paragraph{Controlled retrieval environment.}
Teacher trajectories interact with the same retrieval protocol later exposed
to the student policy.
The corpus is formed by deduplicating the title and content of retrieval
snippets associated with VeriFastScore claims, yielding 1{,}098{,}517 unique
passages.
An E5-base-v2 encoder maps queries and passages to 768-dimensional vectors.
A flat FAISS inner-product index returns the top three passages for every
search.
The retrieved passages are injected into the next environment turn inside an
\texttt{<information>} block.
Using a fixed local index keeps evidence stable across synthesis shards,
training, and controlled evaluation, and ensures that the teacher cannot rely
on capabilities unavailable to the student.

\paragraph{Privileged-teacher principle.}
A strong teacher receives the question, source response, exact ordered gold
claims, and gold binary labels.
This private key is used for three purposes only:
(1) the \texttt{<claims>} array must reproduce the annotated atomic claims
verbatim and in order;
(2) confidence values are calibrated so that ambiguous details are routed to
retrieval; and
(3) \verifytag{} and \answertag{} decisions must match the annotated labels.
The teacher must nevertheless reach those decisions through common knowledge
or retrieved evidence.
It is forbidden to mention the private key, labels, hints, supervision, or any
equivalent source of privileged information in its generated trajectory.

Listing~\ref{lst:teacher-private-prompt} shows the central portion of the
teacher prompt.
The omitted portions give detailed confidence examples, entity-key naming
rules, and the same structural constraints that are checked automatically
after generation.

\begin{lstlisting}[style=policy-prompt,
caption={Privileged-teacher instruction excerpt used during policy demonstration synthesis.},
label={lst:teacher-private-prompt}]
You are an expert long-form factuality verification agent producing a
high-quality reasoning demonstration for a student model to learn from.

[PRIVILEGED ANSWER KEY: never reveal or reference in output]
You are given:
  - the exact atomic claims to extract, and
  - the correct supported or unsupported label for each claim.

The key has only three uses:
  (1) reproduce the claims verbatim and in order in <claims>;
  (2) calibrate confidence for search routing;
  (3) reach the same final labels through genuine evidence reasoning.

Never refer to the key, gold labels, hints, ground truth, privileged
information, or provided supervision in any output token.

Evidence rule:
  supported   = evidence substantiates the specific claim.
  unsupported = evidence contradicts the claim or does not confirm its
                specific date, number, name, or causal link.

Hard constraints:
  - <claims> is parseable JSON with consecutive ids.
  - <search> is the last tag in an action turn.
  - Every search is followed by one <memo> before verification.
  - Claims sharing an entity_key are processed in one group.
  - Every claim receives exactly one <verify>.
  - <answer> contains all claims in order.
  - At most eight searches are issued.
\end{lstlisting}

The corresponding teacher user message is instantiated separately for each
sample:

\begin{lstlisting}[style=policy-prompt,
caption={Teacher user-message template. Braced fields are filled per sample.},
label={lst:teacher-user-prompt}]
Question:
{question}

Response to verify:
{response}

[PRIVATE SUPERVISION: never reveal or reference in any output token]
Decompose the response into EXACTLY these {N} atomic claims, verbatim and
in the given order, and reach EXACTLY these labels in <verify> and <answer>:

1. [supported] {claim_1}
2. [unsupported] {claim_2}
...
N. [supported|unsupported] {claim_N}

All output must read as a genuine investigation. Start with <claims>.
\end{lstlisting}

\paragraph{Five-tag trajectory schema.}
Every retained demonstration follows one compact action language:
\texttt{<claims>} specifies claim decomposition and routing,
\searchtag{} invokes the only external tool,
\memotag{} stores a short evidence summary in the shared context,
\verifytag{} records one claim decision, and
\answertag{} aggregates all claims and labels.
We intentionally omit separate route, recall, and free-form reasoning tags.
The reduced schema exposes the decisions that matter for control while keeping
evidence reuse implicit in the conversation history.

The \texttt{<claims>} array contains four fields per claim.
\texttt{id} is a consecutive integer, \texttt{claim} is the verbatim atomic
claim, \texttt{confidence} is \emph{high}, \emph{medium}, or \emph{low}, and
\texttt{entity\_key} is a normalized central entity or topic.
Claims with the same entity key form a group.
If all members of a group have high confidence, the teacher verifies them
directly in a single batch.
If any member has medium or low confidence, the group issues one focused
search, writes one memo, and then verifies every member, including any
high-confidence members, against the shared context.
This makes search count depend on uncertain entity groups rather than raw claim
count.

\begin{lstlisting}[style=policy-prompt,
caption={Illustrative five-tag policy trajectory. Retrieval observations are generated by the environment.},
label={lst:five-tag-trajectory}]
<claims>
[
  {"id":1,
   "claim":"Einstein won the Nobel Prize in Physics.",
   "confidence":"high",
   "entity_key":"einstein"},
  {"id":2,
   "claim":"Einstein won the Nobel Prize in 1922.",
   "confidence":"medium",
   "entity_key":"einstein"}
]
</claims>

<search>Einstein Nobel Prize in Physics award year</search>

<information>
Doc 1 states that the 1921 Nobel Prize in Physics was awarded to
Albert Einstein and presented in 1922.
</information>

<memo>Einstein received the 1921 Nobel Prize in Physics (Doc 1).</memo>
<verify id=1>supported</verify>
<verify id=2>unsupported</verify>

<answer>{"claims":[
  {"claim":"Einstein won the Nobel Prize in Physics.",
   "label":"supported"},
  {"claim":"Einstein won the Nobel Prize in 1922.",
   "label":"unsupported"}
]}</answer>
\end{lstlisting}

\paragraph{Confidence calibration.}
High confidence is reserved for stable, textbook-level facts that the teacher
can recall specifically.
Medium confidence denotes familiar subject matter with uncertainty about an
exact date, quantity, name, or causal relation.
Low confidence denotes niche, recent, or technical content that requires
retrieval.
The teacher is explicitly instructed not to mark every claim low, which would
teach wasteful search, or every claim high, which would suppress tool use.
Calibration is evaluated indirectly through schema consistency, search
placement, and the empirical distribution of searches rather than by treating
confidence as an independently supervised target.

\paragraph{Entity grouping and evidence reuse.}
Entity keys are lowercase, snake-case identifiers of the central subject.
They are intentionally broader than an individual claim.
For example, claims about Einstein's award, birthplace, and nationality share
the key \texttt{einstein}; a key such as
\texttt{einstein\_birth\_year\_1879} would be rejected as unnecessarily
narrow.
Groups are processed in first-appearance order.
A memo is a one-sentence summary that retains the decisive names, dates,
numbers, and a compact document reference.
Because the memo remains in the dialogue history, later decisions can attend
to it directly.
There is no external memory store and no separate recall operation.

\paragraph{Interactive rollout.}
Generation is paused whenever the teacher completes a \searchtag{} action.
The environment parses the query, returns three passages inside an
\texttt{<information>} turn, and resumes generation with the complete history.
The search closing tag is used as a generation stop condition, ensuring that
the tool action is the final tag of its turn.
At most eight searches and 30 assistant-generation calls are allowed for one
sample.
A generation that reaches its length limit without a complete \answertag{} is
discarded rather than truncated into a nominally valid demonstration.

\paragraph{Quality-control gates.}
Every synthesized sample passes all of the following checks before entering the
training corpus:

\begin{enumerate}
    \item \textbf{Schema validity.}
    The \texttt{<claims>} payload is parseable JSON; claim identifiers are
    consecutive; confidence values belong to the three-level vocabulary; and
    entity keys satisfy the normalized naming rule.

    \item \textbf{Decomposition alignment.}
    The number, order, and stripped text of generated claims match the annotated
    claims exactly.  This prevents teacher paraphrases from changing the unit
    scored by the evaluator.

    \item \textbf{Tool protocol.}
    Search actions occur at turn boundaries, stay within the eight-search
    budget, and are followed by a memo before the corresponding decisions.

    \item \textbf{Decision alignment.}
    Every claim identifier occurs in exactly one \verifytag; its label equals
    the annotation; and the final \answertag{} repeats all claims and labels in
    order.

    \item \textbf{Leakage rejection.}
    Every assistant turn is scanned case-insensitively for references to an
    answer key, gold labels, privileged information, supervision, hints, or
    equivalent phrasing.  A match rejects the complete sample rather than
    editing the offending phrase.

    \item \textbf{Student-view reconstruction.}
    The privileged system and user turns are removed and replaced with the
    deployment prompt.  Assistant actions and retrieval observations are
    preserved verbatim.  The reconstructed conversation is scanned again.

    \item \textbf{Dataset-level audit.}
    Duplicate identifiers and normalized question--response fingerprints are
    removed, and overlap with all development and held-out sets is required to
    be zero.
\end{enumerate}

After quality control and deduplication, the final corpus contains 1{,}447
trajectories.
The multi-stage rejection procedure is deliberately conservative: a malformed,
misaligned, truncated, or potentially leaked sample is dropped rather than
repaired.
Random human spot checks provide an additional semantic audit of tool use and
evidence-label consistency.

\paragraph{Student view.}
The student receives no gold claims or labels.
Its system prompt states the action schema, confidence-conditioned routing
rule, evidence semantics, and search budget.
Its user message contains only the question, long-form response, and an
approximate claim-count cue.
Listing~\ref{lst:student-policy-prompt} provides the complete policy portion of
the prompt used to serialize the final demonstrations.

\begin{lstlisting}[style=policy-prompt,
caption={Student policy prompt stored in the demonstration corpus.},
label={lst:student-policy-prompt}]
You are a long-form factuality verification agent.
You will receive a question and a model's response. Verify the response by:
decomposing it into atomic claims with confidence calibration, grouping by
entity, and verifying each group with search when needed.

Process:
  1. Emit <claims> as a JSON array. Each object has:
       "id":          int 1..N
       "claim":       atomic claim text, verbatim from the response
       "confidence":  "high" | "medium" | "low"
       "entity_key":  lowercase snake_case central entity

  2. Process each unique entity_key in first-appearance order:
     - If all claims in the group are high confidence, output batched
       <verify> tags only.
     - If any claim is medium or low confidence, output one <search>
       covering the group, wait for <information>, write one <memo>,
       and then verify all claims in the group.

  3. Finish with:
       <answer>{"claims":[
         {"claim":"...", "label":"supported|unsupported"}, ...
       ]}</answer>

Rules:
  - Search only when a group contains a medium or low confidence claim.
  - After every useful search, write a memo before verification.
  - Process all claims sharing an entity_key in one group turn.
  - "unsupported" includes cases where evidence does not confirm the
    specific date, number, name, or causal link.
  - Use at most eight searches.
\end{lstlisting}

\begin{lstlisting}[style=policy-prompt,
caption={Student user-message and tool-description template.},
label={lst:student-user-prompt}]
Question:
{question}

Model Response:
{response}

(This response contains approximately {num_claims} verifiable claims.
Verify them as described above.)

You have access to one tool:
  <search>query</search>
The retrieval engine replies inside <information>...</information>.

After every useful search, write:
  <memo>one-line distilled evidence (Doc X)</memo>

Output the final aggregate in:
  <answer>{"claims":[...]}</answer>
\end{lstlisting}

\paragraph{Final corpus statistics.}
The 1{,}447 retained trajectories contain 19{,}334 claim decisions.
Of these, 70.4\% are \support{} and 29.6\% are \unsupport.
The mean trajectory issues 2.24 searches.
Mean generated trajectory length, including injected retrieval observations
after the initial prompt, is 9.16 thousand characters.
Character length is reported because it is independent of the tokenizer used
for each backbone.
The fact that search count is much smaller than mean claim count reflects both
direct verification of high-confidence groups and reuse of one retrieval
result across related claims.

\subsection{Preference-Pair Construction}
\label{sec:appendix-preference-data}

\paragraph{Motivation and candidate rollouts.}
Policy demonstrations teach decomposition, routing, search, memo formation,
and factuality decisions jointly.
Preference data are constructed only after this policy has converged, and are
used to correct residual decision errors.
We generate policy rollouts on a candidate pool that is disjoint from the
development and held-out evaluation sets.
Only structurally valid trajectories with at least one incorrect verdict are
considered.
We further require the existing context to contain sufficient evidence for the
gold decision so that the selected case represents a decision failure, not a
missing-search failure.

\paragraph{Same-trajectory counterfactuals.}
For each eligible trajectory, the original model completion becomes the
rejected member of a preference pair.
The chosen member is produced by copying the complete trajectory and correcting
only erroneous labels in two synchronized locations:
the corresponding \verifytag{} content and the matching \texttt{label} value
inside \answertag.
Claim text, identifiers, confidence, entity keys, group order, search queries,
retrieval observations, memos, turn boundaries, and all other tokens are kept
fixed.
This construction turns each pair into a controlled counterfactual in which
the factuality decision is the only changed variable.

\begin{lstlisting}[style=policy-prompt,
caption={Schematic same-trajectory preference pair. All omitted context is byte-identical.},
label={lst:dpo-pair-example}]
Shared trajectory prefix:
  <claims>...</claims>
  <search>...</search>
  <information>...</information>
  <memo>...</memo>

Rejected:
  <verify id=7>supported</verify>
  ...
  {"claim":"...", "label":"supported"}

Chosen:
  <verify id=7>unsupported</verify>
  ...
  {"claim":"...", "label":"unsupported"}
\end{lstlisting}

Conventional preference construction from independently sampled completions
would allow the chosen and rejected members to differ in search count,
retrieved documents, decomposition style, memo content, length, and factuality
labels.
Those differences make it difficult to identify which behavior the optimizer
is rewarding.
In a same-trajectory pair, identical tokens cancel in the chosen-versus-rejected
comparison, while the decision-token mask further excludes every unchanged
position from the explicit sequence score.
The full evidence remains visible as context at the active label positions.

\paragraph{Bidirectional corrections.}
A corpus containing only false-support corrections would lower \fsr but could
teach a degenerate policy that predicts \unsupport{} for every uncertain claim.
We therefore retain both correction directions.
There are 405 false-support-to-unsupported pairs and 389
false-unsupported-to-supported pairs, giving 794 pairs in total.
The near-balanced direction count allows the preference signal to target
specific evidence-conditioned errors rather than shift the global class prior.
The chosen completions contain 15{,}686 claim decisions, of which 61.7\% are
\support{} and 38.3\% are \unsupport.
They average 3.69 searches and 13.80 thousand characters per trajectory.

\paragraph{Preference-pair validation.}
Each pair is admitted only after the following invariants are checked:

\begin{enumerate}
    \item The rejected rollout is structurally valid and its erroneous claim
    identifiers are matched to gold annotations.
    \item Every corrected \verifytag{} label is synchronized with the
    corresponding \answertag{} label.
    \item Chosen labels match the gold decisions, while the rejected completion
    retains at least one identified false-support or false-unsupported error.
    \item After replacing decision values with placeholders, chosen and rejected
    trajectories are identical.  This checks the same-trajectory property.
    \item Both correction directions are retained during balancing; no
    unsupported-only label policy can satisfy all pairs.
    \item Duplicate pairs and question--response fingerprint overlap with
    development or held-out evaluation pools are rejected.
\end{enumerate}

\paragraph{Relationship to the training loss.}
The pair construction and decision-focused loss enforce the same locality at
two different levels.
Data construction controls what differs between $y^+$ and $y^-$; the loss mask
controls which token log probabilities contribute to the optimization
objective.
The active positions are only label values inside \verifytag{} and
\answertag.
Search and evidence tokens remain in the causal prefix, so a decision can still
depend on retrieved information, but they receive no direct preference reward.

\end{document}